# An OLAC Extension for Dravidian Languages

*Dr. B Prabhulla Chandran Pillai,*
*Catholicate College,*
*(Affiliated to Mahatma Gandhi University, Kottayam),*
*Pathanamthitta, Kerala, India[1]*

## Abstract

OLAC was founded in 2000 for creating online databases of language resources. This paper intends to review the bottom-up distributed character of the project and proposes an extension of the architecture for Dravidian languages. An ontological structure is considered for effective natural language processing (NLP) and its advantages over statistical methods are reviewed

## 1 Introduction

Though the compilation of morphological and phonological rules are effectively done using finite-state acceptors and transducers, (Mehryar Mohri, Fernando Pereira, Michael Riley, 1996) when it comes to Dravidian languages a semantic counter revolution is very much needed and it should be based on an ontological structure that also considers the common sense view of human brain. It has already been shown that (W. S. Saba, 2008), this is true in the case with non phonetic languages like English.

In this paper, the author tries to provide a technically feasible methodology for the preservation of Dravidian Languages based on an OLAC Extension and a semantic ontological structure. Further, using variables and quantification we can improvise the whole concept of Ontology. Fred Sommers (1963) suggested the idea of strongly typed ontology that is quite evident in the spoken language, *where two objects x and y are considered to be of the same type iff the set of monadic predicates that are significantly (that is, truly or falsely but not absurdly) predicable of x is equivalent to the set of predicates that are significantly predicable of y.*

The agreement to this idea makes the phonetic languages like Dravidian languages more computationally analyzable languages.

The knowledge- based approaches in the ordinary natural language processing (NLP) using quantitative (say statistical, corpus-based or machine learning) tools are not fully fool proof.

## 2. Review

The preservation and nourishing of the language is viable only if the mode of accessing is akin to the one illustrated in the figure (Fig. 1) (Steven BirdGary Simons and Chu-Ren Huang, 2001)

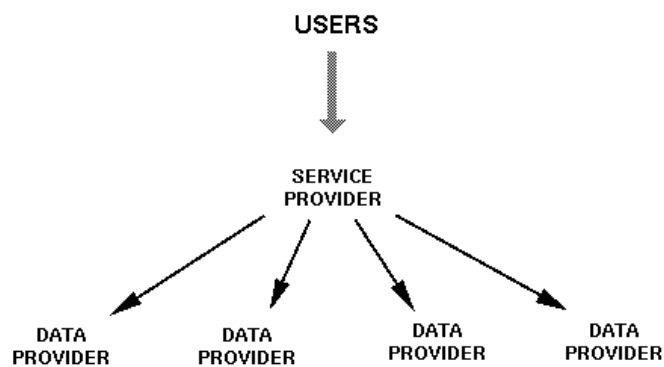

The Open Language Archives Community (OLAC) was formulated to have a federated language archives based on the *Open Archives Initiative* and the *Dublin Core Metadata Initiative.*

---

[1] Email: bprabhullachandran@gmail.com

The project is mostly pivoted around the goal that enables researchers and linguists to document their resources, tools and data. In order to avoid the complexities in finding the right information, the project employed the OAI *Shared Metadata Set* (Dublin Core), which facilitated interoperability across all repositories that are involved in the OAI and the OAI *Metadata Harvesting Protocol*, which helps the user to query a repository using HTTP requests. Though OAI archives have a submission procedure, a long-term storage system, and a retrieval mechanism, it is not effective for preserving the classical documents which are very vital when it comes to Dravidian languages like Sanskrit. So is the case with the folklore languages that are strongly connected to this.

Even though the project follows a bottom-up distributed character of the web, it id quite effective since it has a *structured nature of a centralized database*. What interests more to the Dravidian language linguists is that it can support metadata standards in addition to the Dublin Core. The user (using Ethnologue database) has also access to specific information for each language; viz. *a unique three-letter code, the country where this language is spoken, alternative names, dialects, language classification, comments, and references to the SIL bibliography*.

Here we summarize the Meta information normally carried by the OLAC entry:

Title:
Creator:
Subject:
Subject.language:
Description:
Publisher:
Contributor:
Date:
Type:
Type.linguistic:
Type.functionality:
Format:
Format.cpu:
Format.encoding:
Format.markup:
Format.os:
Format.sourcecode:
Source:
Language:

Though language identification and supporting multilingual resources are few hurdles in the way, OLAC still stands out as a potential candidate for language preservation (and promotion).

We can find that the MT (*Machine Translation*) system is a unidirectional system which has the ability to translate language A to language B. And for handling the bidirectional situation, where a system translates in both directions between the languages, say between X and Y, as illustrated below:

```
pair><Subject.language code= X/>
<Language code= Y/></pair>
<pair><Subject.language code= Y/>
<Language code= X/></pair>
```

And a fully bidirectional lexicon specified for the users of either language would be described in the same way as a bidirectional MT system as shown below:

```
<Language code= X/>
<Language code= Y/>
<Subject.language code= X/>
<Subject.language code= Y/>
```

## 3. Analysis and Suggestions

The disappointments with purely engineering methodologies were discussed briefly in the introduction. And these eventually led to the construction of very large knowledge bases for NLP (Lenat and Ghua, 1990). Thus with a sound linguistic and ontological underpinnings, a good extension of OLAC can be done for the preservation (and promotion) of Dravidian languages.

The root cause for the development is that most of the documents still exist in manuscript format. And when it comes to languages like Sanskrit, the resources are mostly in 'copylefted state'. NLP can help in converting this to digitally readable format and using MT technique the languages can be popularised. These two points are elucidated further in the rest of this paper.

The whole scenario is practical since we are in a Platonic universe which admits the existence of anything that we can discuss using our language. Cocchiarella (1996) went one more step ahead by arguing that the language can speak about things that may have existed in the past or may exist in the future. It is interesting to note that, once we endorse these ideas the processing of Dravidian languages (especially Dravidian literature) becomes a task that can be 'quantised'.

Now, as suggested by Saba, let's take a very simple case.

a. *Olga is an old dancer.*
b. *Olga is an elderly teacher.*

This can be represented quantitatively using mathematical expressions. But we are after ontological representations. So, we can have show this as (Saba, 2008)

$$⟦Olga\ is\ a\ beautiful\ dancer⟧$$
$$⟹ (\exists^1 Olga :: \text{Human})(\exists a :: \text{Activity})$$
$$(\text{DANCING}(a) \wedge \text{AGENT}(a, Olga :: \text{Human}) \wedge$$
$$(\text{BEAUTIFUL}(a :: \text{Entity}) \vee \text{BEAUTIFUL}(Olga :: \text{Entity}))$$

$$⟦Olga\ is\ a\ beautiful\ dancer⟧$$
$$⟹ (\exists^1 Olga :: \text{Human})(\exists a :: \text{Activity})$$
$$(\text{DANCING}(a) \wedge \text{AGENT}(a, Olga)$$
$$\wedge (\text{BEAUTIFUL}(a) \vee \text{BEAUTIFUL}(Olga)))$$

$$⟦Olga\ is\ an\ elderly\ teacher⟧$$
$$⟹ (\exists^1 Olga :: \text{Human})(\exists a :: \text{Activity})$$
$$(\text{TEACHING}(a) \wedge \text{AGENT}(a, Olga :: \text{Human}) \wedge$$
$$(\text{ELDERLY}(a :: \text{Human}) \vee \text{ELDERLY}(Olga :: \text{Human})))$$

$$⟦Olga\ is\ an\ elderly\ teacher⟧$$
$$⟹ (\exists^1 Olga :: \text{Human})(\exists a :: \text{Activity})$$
$$(\text{TEACHING}(a) \wedge \text{AGENT}(a, Olga :: \text{Human})$$
$$\wedge (\text{ELDERLY}(a :: (\text{Human} \bullet \text{Activity}))$$
$$\vee \text{ELDERLY}(Olga :: \text{Human}))$$
$$⟹ (\exists^1 Olga :: \text{Human})(\exists a :: \text{Activity})(\text{TEACHING}(a)$$
$$\wedge \text{AGENT}(a, Olga) \wedge (\bot \vee \text{ELDERLY}(Olga))$$
$$⟹ (\exists^1 Olga :: \text{Human})(\exists a :: \text{Activity})$$
$$(\text{TEACHING}(a) \wedge \text{AGENT}(a, Olga) \wedge \text{ELDERLY}(Olga))$$

Since the Dravidian languages are phonetic in nature the processing becomes easier as the presentations are simpler. Just as in the final step, we had 'Olga is a beautiful dancer' as *Olga is the agent of some dancing Activity, and either Olga is*

*BEAUTIFUL or her DANCING (or, of course, both).* And for the second statement as: *there is a unique object named Olga, an object that must be of type Human, and an object a of type Activity, such that a is a teaching activity, Olga is the agent of the activity, and such that elderly is true of Olga.*

NLTK (Natural Language Processing Kit) can handle these using threading and combination of different rule types as shown (Edward Loper and Steven Bird, 2002) below:

```
cascade = [
ChunkRule('<DT><NN.>*<VB.><NN.>'),
ChunkRule('<DT><VB.><NN.>'),
ChunkRule('<.>'),
UnChunkRule('<JN><VB.>*<CC>|<MD>|<R.>.'),
UnChunkRule('<.|\\.>|.'),
MergeRule('<NN.>*<DT_JJ>*<CD>',
'<NN.>*<DT_JJ>*<CD>'),
SplitRule('<NN.>,<DT_JJ>')
]
```

The Weighted Transducers' (Mehryar Mohri, Fernando Pereira, Michael Riley, 1996) method will be further incorporated to improvise the overall efficiency as shown below:

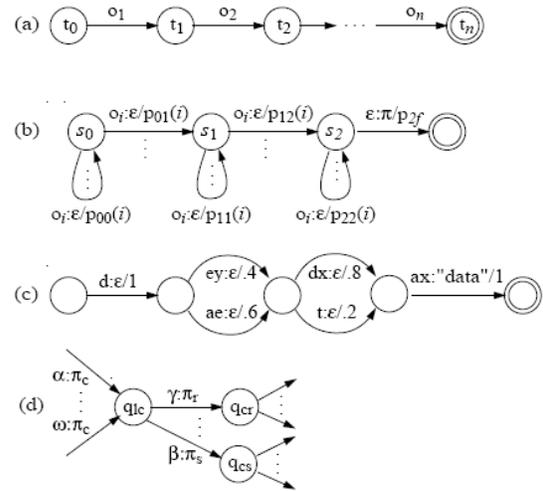

## 4. Conclusion and Remarks

The existing structure of OLAC can easily be extended (by keeping the architecture structure as such) to incorporate the *copylefted* contents from the Dravidian literature branch. By opting for an effective cloud (or parallel) computing architecture the processing can be made effective and faster. For best results, we need to follow a semantic and ontological structure based on commonsense observations rather than purely engineering methodologies.

## 5. References

Montague, R. (1970), English as a Formal Language, In R. Thomasson (Ed.), *Formal Philosophy – Selected Papers of Richard Montague*, New Haven, Yale University Press.

Montague, R. (1960), On the Nature of certain Philosophical Entities, *The Monist*, 53, pp. 159-194.

Ng, Tou H. and Zelle, J. (1997), Corpus-based Approaches to Semantic Interpretation in Natural language, *AI Magazine*, Winter 1997, pp. 45-64.

Partee, B. (2007), Compositionality and Coercion in Semantics – the Dynamics of Adjective Meanings, In G. Bouma et. al. (Eds.), *Cognitive Foundations of Interpretation, Amsterdam*: Royal Netherlands Academy of Arts and Sciences, pp. 145-161.

Partee, B. H. and M. Rooth. (1983), Generalized Conjunction and Type Ambiguity. In R. Bauerle, C. Schwartze, and A. von Stechow (eds.), *Meaning, Use, & Interpretation of Language*. Berlin: Walter de Gruyter, pp. 361-383.

Rais-Ghasem, M. and Coriveaue, J.-P. (1998), Exemplar-Based Sense Modulation, *In Proceedings of COLING-ACL '98 Workshop on the Computational Treatment of Nominals*.

Saba, W. (2007), Language, logic and ontology: Uncovering the structure of commonsense knowledge, *International Journal of Human-Machine Studies*, 65(7), pp. 610-623.

*Encyclopedia of Linguistics*. Oxford University Press.

DCMI. 2000. Dublin Core qualifiers. http://dublincore.org/documents/2000/07/11/dcmes-qualifiers/.

Barbara F. Grimes, editor. 2000. *Ethnologue: Languages of the World*. Dallas: Summer Institute of Linguistics, 14th edition

Carl Lagoze and Herbert Van de Sompel. 2001. The Open Archives Initiative: Building a low-barrier interoperability framework.

Ronald Cole, editor. 1997. *Survey of the State of the Art in Human Language Technology*. Studies in Natural Language Processing. Cambridge University Press.

DCMI. 1999. Dublin Core Metadata Element Set, version 1.1: Reference description. http://dublincore.org/documents/1999/07/02/dces/ .

[Carlsson & Beldiceanu 2002] Carlsson, M., and Beldiceanu, N. 2002. Arc-consistency for a chain of lexicographic ordering constraints. Technical report T2002-18, Swedish Institute of Computer Science.

Demassey, S.; Pesant, G.; and Rousseau, L. 2006. A cost-regular based hybrid column generation approach. *Constraints* 11(4):315– 333.

Golden, K., and Pang, W. 2003. Constraint reasoning over strings. In Rossi, F., ed., *Proc. of 9th Int. Conf. on Principles and Practice of Constraint Programming (CP2003)*, 377–391.

Seigel, E. (1976), *Capturing the Adjective*, Ph.D. dissertation

Steven Bird and Gary Simons. 2001. The OLAC metadata set and controlled vocabularies. In *Proceedings of ACL/EACL Workshop on Sharing Tools and Resources for Research and Education*. http://arXiv.org/abs/cs/0105030.